\documentclass[letterpaper, 10 pt, conference]{ieeeconf}  

\IEEEoverridecommandlockouts                              

\usepackage{graphics} 
\usepackage{epsfig} 
\usepackage{mathptmx} 
\usepackage{times} 
\usepackage{amsmath} 
\usepackage{amssymb}  
\usepackage{xcolor}
\usepackage{booktabs}
\usepackage{array}
\usepackage[normalem]{ulem}
\usepackage[noadjust]{cite}
\usepackage{footnote}
\usepackage{hyperref}
\useunder{\uline}{\ul}{}

\title{\LARGE \bf
An Integrated Actuation-Perception Framework for Robotic\\ Leaf Retrieval: Detection, Localization, and Cutting
}

\author{Merrick Campbell, Amel Dechemi, and Konstantinos Karydis
\thanks{The authors are with the Dept. of Electrical and Computer Eng., Univ. of California, Riverside, 900 University Avenue, Riverside, CA 92521, USA. Email: {\tt\footnotesize\{mcamp077, adech003, karydis\}@ucr.edu}. 
We gratefully acknowledge the support of USDA-NIFA under grant \# 2021-67022-33453 and a UC MRPI Award. 
Any opinions, findings, and conclusions or recommendations expressed in this material are those of the authors and do not necessarily reflect the views of the funding agencies.
}}

\begin{document}

\maketitle
\thispagestyle{empty}
\pagestyle{empty}

\begin{abstract}

Contemporary robots in precision agriculture focus primarily on automated harvesting or remote sensing to monitor crop health. Comparatively less work has been performed with respect to collecting physical leaf samples in the field and retaining them for further analysis. Typically, orchard growers manually collect sample leaves and utilize them for stem water potential measurements to analyze tree health and determine irrigation routines. While this technique benefits orchard management, the process of collecting, assessing, and interpreting measurements requires significant human labor and often leads to infrequent sampling. Automated sampling can provide highly accurate and timely information to growers. The first step in such automated in-situ leaf analysis is identifying and cutting a leaf from a tree. This retrieval process requires new methods for actuation and perception. We present a technique for detecting and localizing candidate leaves using point cloud data from a depth camera. This technique is tested on both indoor and outdoor point clouds from avocado trees. We then use a custom-built leaf-cutting end-effector on a 6-DOF robotic arm to test the proposed detection and localization technique by cutting leaves from an avocado tree. Experimental testing with a real avocado tree demonstrates our proposed approach can enable our mobile manipulator and custom end-effector system to successfully detect, localize, and cut leaves.

\end{abstract}

\section{Introduction}

Precision agriculture is a farming practice that utilizes sensor networks to help improve the use of agronomic inputs (e.g., water, fertilizers, pesticides)~\cite{ZHANG2002113}. Robotics research in precision agriculture has largely focused on remote sensing via ground or aerial robots (e.g.,~\cite{maes2019perspectives,radoglou2020compilation,kim2019unmanned}). Besides remote sensing, an increasing number of works has begun addressing interactions with the crop. Such works consider primarily robotic harvesting in both row (e.g., corn and soybean) and tree crops (e.g., citrus and avocado). For example, robots have been deployed to pick peppers, apples, citrus, and tomatoes by wrapping the fruit and twisting it off the stem with either a soft gripper~\cite{lehnert2017autonomous,hohimer2019design,chowdhary2019soft}, rigid gripper~\cite{mehta2014vision,mehta2016robust,de2011design,davidson2017dual,davidson2016proof,nguyen2013task,uppalapati2020berry}, or vacuum~\cite{schupp2011preliminary,baeten2008autonomous,zhang2020system}. Some robots can pick strawberries, cucumbers, citrus, and peppers by cutting the stem~\cite{hayashi2010evaluation,van2009optimal,van2003field,aloisio2012next,arad2020development,r2018research}.

This paper focuses on interaction with tree crops and addresses a conceptually-related yet less explored topic compared to robotic harvesting: \emph{robotic leaf sampling}. Leaf sampling is important in agriculture since remote sensing typically provides field-level information without sufficient resolution to accurately diagnose problems. Agronomists utilize specialized instruments that can be difficult to transport to the field and thus rely upon sample retrieval for later lab analysis. While this has been mostly a manual process to date, some work has been performed using aerial and ground robots. Mueller-Sim et al. demonstrated a robotic platform for rapid phenotyping and capable of manipulating leaves for in-situ measurements~\cite{mueller2017robotanist,abel2018fieldrobotic}. Orol et al. developed a tele-operated aerial robot for cutting and collecting leaves from trees~\cite{VijayICUAS2017}. Ahlin et al. presented an algorithm for selecting and grasping tree leaves using a robotic arm~\cite{ahlin2016autonomous}. The latter work demonstrates a high level of control using monoscopic depth analysis (MDA) and image-based visual servoing, but focuses on grasping and pulling the leaf instead of cleanly cutting the stem of the leaf, which is the focus of our work.

\begin{figure}[!t]
\vspace{6pt}
    \centering
    \includegraphics[trim={0cm 5cm 0cm 3cm },clip,width=0.95\linewidth]{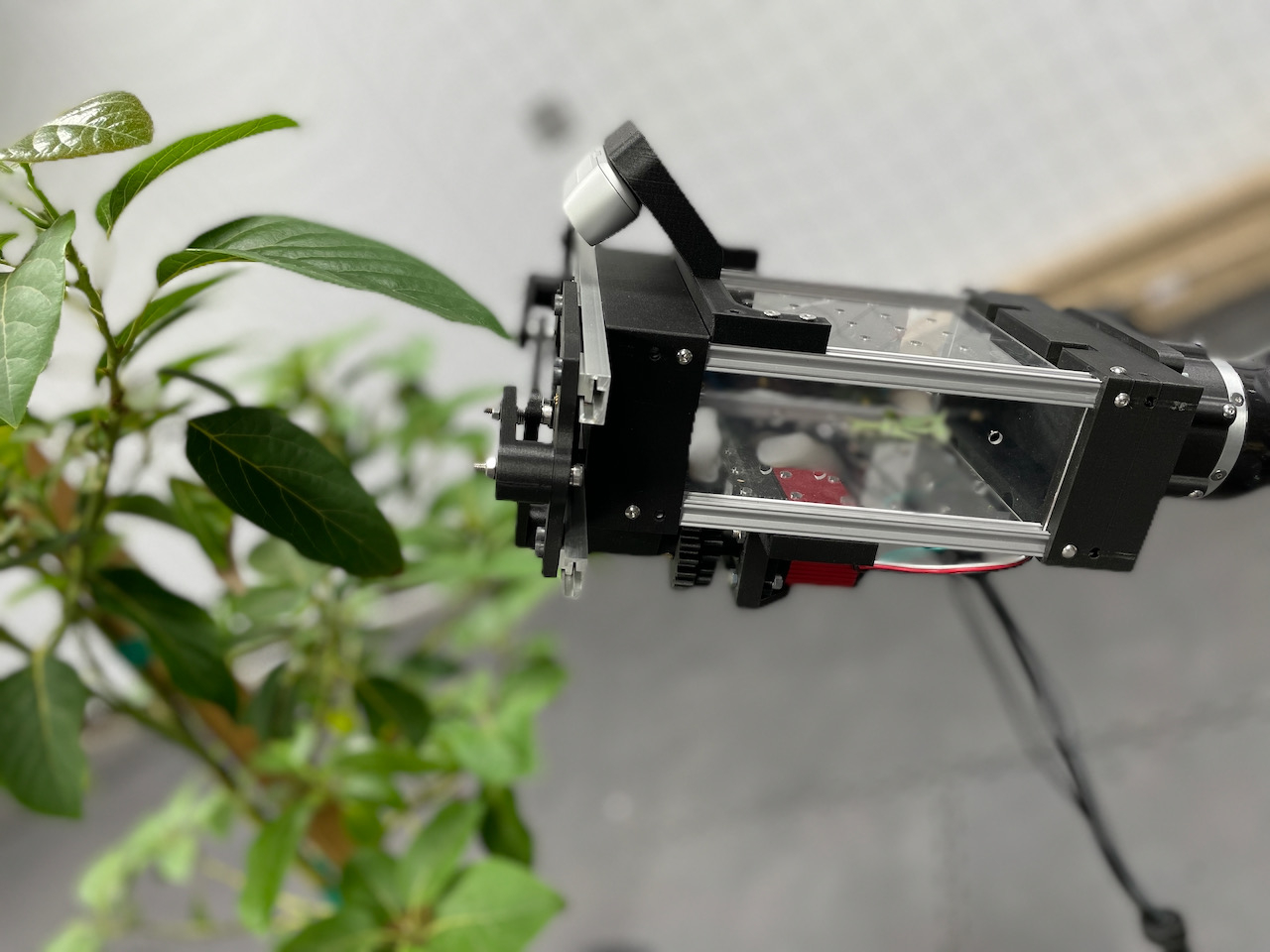}
    \vspace{-6pt}
    \caption{We develop a custom-built end-effector attached to an off-the-shelf 6-DOF robotic arm and a visual perception algorithm to detect, localize and cut leaves at their stem.  (The supplementary video demonstrates the end-effector's operation and overall system testing.)
    }
    \label{fig:eefonarm}
    \vspace{-18pt}
\end{figure}

Our work is motivated by the need to perform leaf water potential measurements, an important process performed by agronomists to estimate tree stress levels and hence optimize irrigation patterns~\cite{UCANRpressureIrrigation}. %
A leaf cut at its stem is placed inside a pressurized chamber instrument with its cut end exposed, and then the pressure at which water begins to escape from the cut stem is used to determine the leaf water potential~\cite{scholander1965sap,tyree1972measurement}. Agronomists use this measurement as a proxy for tree stress levels to optimize irrigation patterns. 
Though effective, these instruments can be tedious and potentially dangerous to operate.\footnote{\url{www.pmsinstrument.com/maintenance/safety/}}
As a result, a single tree is often used to quantify the health of the entire orchard leading to infrequent measurements and undersampled regions. Enabling robotic leaf sampling (this paper's focus) for future use in robotic leaf water potential analysis can help improve measurement coverage and frequency while reducing human fatigue, and risk of bodily injury. Our work joins a growing body of works on robotic means for monitoring crop health and improving irrigation management practices~\cite{RAPIDBookChapter, CARPINCASE2108B}.

Compared to existing robotic leaf sampling methods~\cite{mueller2017robotanist, abel2018fieldrobotic, VijayICUAS2017, ahlin2016autonomous} and harvesting systems that cut the stem of a fruit/vegetable~\cite{hayashi2010evaluation, van2009optimal, van2003field, aloisio2012next, arad2020development, r2018research}, we are interested in performing clean cuts at leaves' stems and retaining leaves for stem water potential analysis.
%
%
As with related works, we also incorporate a visual perception component (to identify and localize a leaf) and an actuation component (to move the end-effector toward the leaf, and then cut it). Collecting a leaf sample from a tree presents unique challenges in perception and actuation, distinct from robotic fruit harvesting.
Similarly, finding a motion plan to retrieve a physical sample needs to account for the presence of other leaves and branches which can also interfere with the extraction process. Yet, identifying a leaf sample involves not only segmenting the canopy, but also selecting an unblemished leaf suitable for stem water potential analysis~\cite{UCANRpressureIrrigation}.



To this end, we propose a leaf-cutting end-effector combined with a visual perception system that detects the center of a leaf and estimates its 6D pose (Fig.~\ref{fig:eefonarm}). The end-effector can cut and capture leaves of several common tree crops, such as avocado, clementine, grapefruit, and lemon. Unlike the MDA approach~\cite{ahlin2016autonomous}, we use a depth camera and a 3D point cloud to identify the centroid of the leaf and then estimate its 6D pose. This paper outlines our perception and actuation process to detect, localize, and cut leaves at their stem while retaining them, to enable future automated leaf water potential analysis in tree crops.

\section{Related Works}
Development of harvesting end-effectors is an active area of research due to the wide variety of crops. While there are some commonalities across approaches, differences in size, weight, shape, texture, and firmness of specialty crops have led to unique solutions. Apples and citrus require a specific motion to grasp, twist, and pull from the tree without damage~\cite{bu2020experimental,aloisio2012next}. Bell peppers and cucumbers can be directly cut and harvested~\cite{lee2019vision,arad2020development,van2009optimal,van2003field}. More delicate crops like strawberries call for manipulators with force feedback and flexible pneumatic actuators~\cite{Simonton1991,hayashi2010evaluation,xiong2020autonomous,bao2015flexible}. 
Despite their unique applications, harvesting end-effectors generally have three primary components: the gripping mechanism (mechanical, pneumatic or  hybrid), the removal mechanism (mechanical or electrical), and the sensing modality (monocular camera, stereo camera, time-of-flight)~\cite{morar2020robotic,r2018research}.



At the same time, there has been development of perception techniques to monitor
crop growth~\cite{ZHU201628,sadeghi2017automated}, help prevent disease through early detection~\cite{akram2017towards,8706936}, 
assist with quality control~\cite{su2018potato,JAHNS200117}, and help automate harvesting~\cite{bu2020experimental}. 
Success of these tasks depends on the visual perception subsystem's ability to provide precise and accurate information about the target crop and relevant environmental context~\cite{2021}, including 
segmentation and localization of targets of interest. 
%
Most approaches have focused on fruit/vegetable targets by harnessing distinct colors and/or shapes~\cite{ahlin2016autonomous,FU2020105687,chen2017counting,NGUYEN201633,7942724}. 

In this paper we are targeting identification and pose estimation of individual tree-crop leaves. This presents similar yet unique challenges compared to fruit (and broader canopy) identification. 
Instead of filtering out the leaves to focus on the fruit, our objective is to retain the leaves and segment the tree canopy further to obtain individual leaf poses. 
Leaf segmentation has been considered in current research using both classical computer vision tools~\cite{Chen_2019_CVPR_Workshops,MIAO2021106310,ELNASHEF201951} as well as machine learning~\cite{Guo2021LeafMaskTG, 9025429, Scharr2015LeafSI}. However, classical methods are prone to changes in the environment, such as light, occlusions or overlapping surfaces, whereas learning-based methods require large training datasets and may still generalize poorly as environmental factors vary~\cite{app11010228}.  

Furthermore, these techniques have rarely been employed online on onboard computers as part of a robotic manipulation system to identify, localize and physically cut the leaf. 
Although a leaf's 3D position can be readily obtained, it is not sufficient to successfully accomplish the task as orientation plays an important role as to how a robotic arm approaches the leaf to cut it. 
Thus, obtaining at least an estimate of the 6D pose (position and orientation) is critical. 
Traditional 6D pose estimation approaches usually perform local keypoint detection and feature matching, and then a RANSAC-based PnP algorithm on the established 3D-to-2D correspondences to estimate the pose of an object ~\cite{Michel2017GlobalHG, Brachmann2016UncertaintyDriven6P}. Still, they typically fail to perform with heavily occluded and poorly textured objects. On the other hand, learning-based methods use a deep neural network (DNN) to obtain the correspondences between 3D objects points and their 2D image projections~\cite{Hu2019SegmentationDriven6O,Hu2020SingleStage6O, Park2019Pix2PosePC}. Use of synthetic data generators~\cite{Giuffrida2017ARIGANSA, Zhu2018DataAU} can relieve in part the challenge of acquiring large labeled datasets; however, it requires realistic models that take into account the variations of the detected object e.g., shape, size, orientation or curvature which can be hard to develop.


Our developed end-effector focuses on the actuation and perception techniques to cut and retain a leaf. This task has received much less attention by existing robotic harvesting/leaf sampling technology yet an important aspect toward enabling future robotic leaf water potential measurements.

\section{Technical Approach}

\begin{figure}[!t]
\vspace{6pt}
    \centering
    \includegraphics[trim={0 0 0 0 },clip,width=0.975\linewidth]{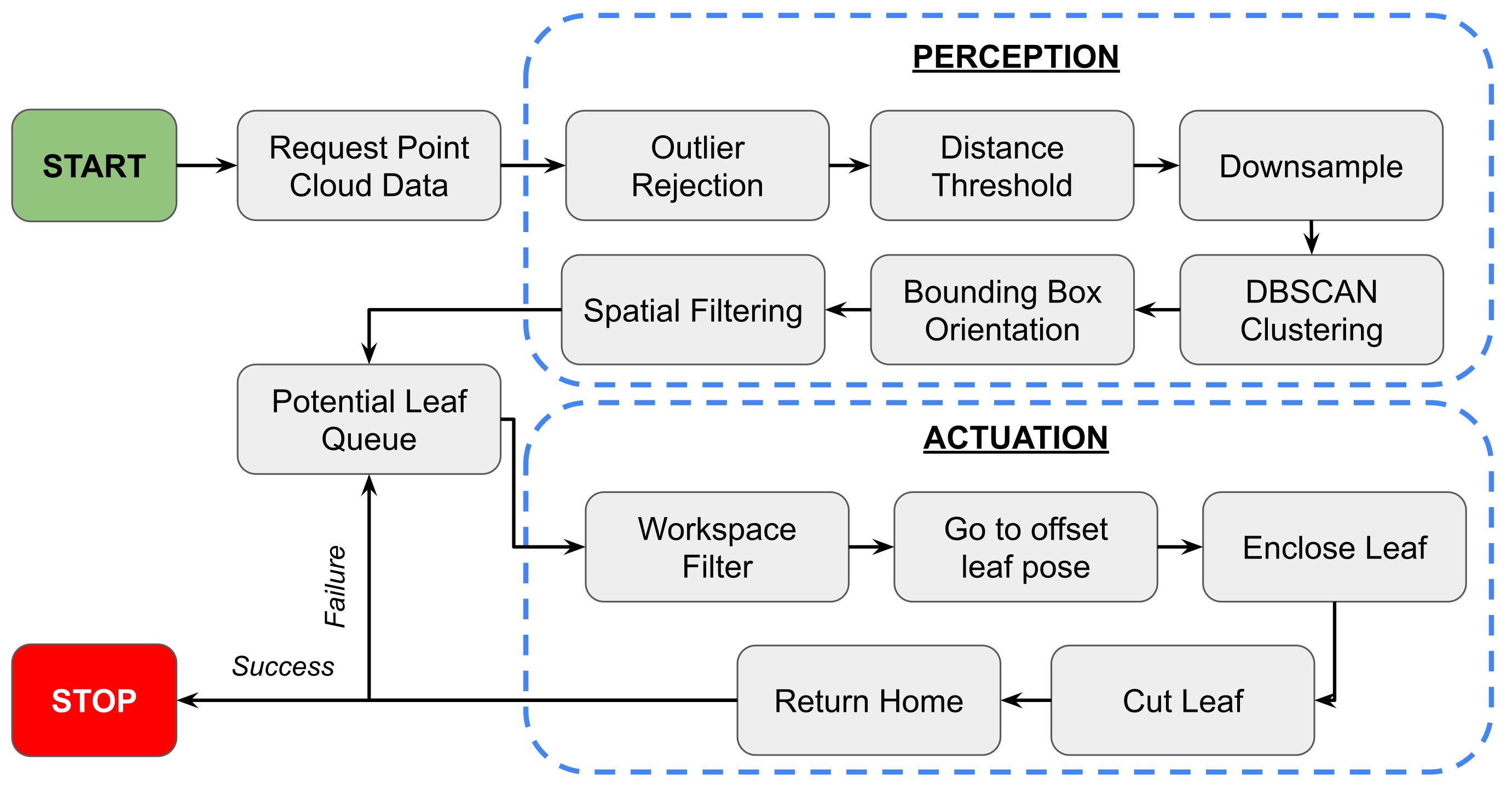}
    \vspace{-10pt}
    \caption{Our approach jointly considers perception and actuation. The perception module processes point cloud data to segment leaves and deposit leaf candidates into a queue. Candidate leaves are then passed to the robot arm controller to actuate the end-effector. If a cut is successful, the routine ends. If unsuccessful, the arm controller requests the next leaf in the queue.
    }
    \label{fig:systemdiagram}
    \vspace{-18pt}
\end{figure}

Picking a leaf requires two key components: actuation and perception. For actuation, we design a custom-built leaf-cutting end-effector (Section~\ref{sec:ta_leaf_cutting}) and retrofit it on a mobile manipulation base platform (Kinova Gen-2 six degree of freedom [6-DOF] robot arm mounted on a Clearpath Robotics Husky wheeled robot). For perception, we utilize point cloud data from a depth camera (Intel RealSense D435i) for the leaf detection and localization algorithm developed herein (Section~\ref{sec:ta_leaf_perception}). The point cloud data is processed using Open3D~\cite{Zhou2018} running on an Intel i7-10710U CPU, without any additional GPU acceleration. Figure~\ref{fig:systemdiagram} highlights how our contributions interact in a leaf-cutting system, which is further evaluated in Section~\ref{sec:exp_methods}.

Identified and segmented leaves serve as target for the arm to move and align the end-effector along a viable leaf (to be defined in Section~\ref{sec:exp_methods}), at an offset position from the center of the leaf. The offset distance is equivalent to the length of the leaf. Once at the offset position, the arm moves linearly toward the leaf to capture it. When the leaf is enclosed, the end-effector cuts the leaf. Then, the arm returns home.

\subsection{Actuation} 
\label{sec:ta_leaf_cutting}

\begin{figure}[!t]
\vspace{6pt}
    \centering
    \includegraphics[trim={0 0 0 0 },clip,width=0.90\linewidth]{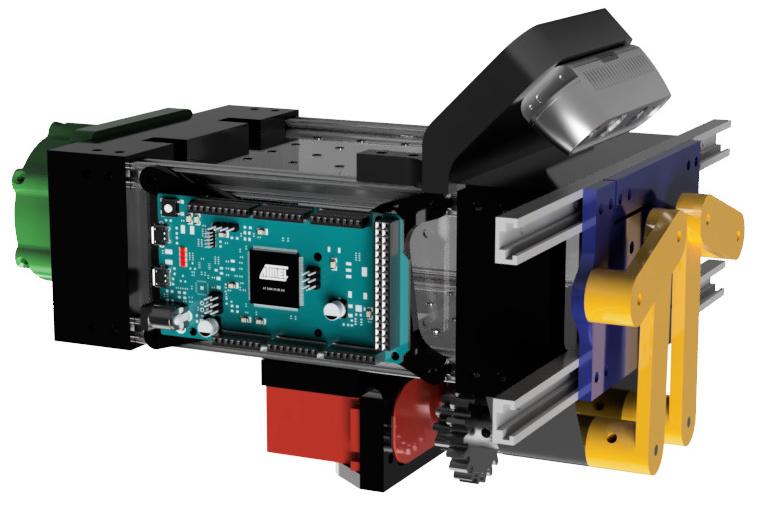}
    \vspace{-9pt}
    \caption{The end-effector contains the components necessary to cut a leaf from a tree. The servo motor (red) actuates a double four-bar mechanism (yellow) that closes a set of gates (blue) with a razor blade to cut and capture a leaf. An Intel RealSense camera D435i is mounted on the top of the end-effector for perception. A microcontroller is mounted on the arm for controlling the motor. This end-effector can be mounted to a robotic arm using an adaptor plate (green). (Figure best viewed in color.)
    }
    \label{fig:eef_cad}
    \vspace{-18pt}
\end{figure}

The stem-cutting end-effector developed herein utilizes two 4-bar linkages to actuate a set of sliding gates, one of which contains a razor blade to remove the leaf from the tree (Fig.~\ref{fig:eef_cad}). The gates also help retain the leaf within the end-effector's chamber after removal from the tree. These 4-bar mechanisms are connected via a gear train to achieve synchronized motion.  A low-cost, high-torque R/C servo (FEETECH FT5335M) drives the gear train while being amenable to position control. An Arduino Due microcontroller controls the servo motor and receives serial commands from a ROS control node. A breakout board connected to the Arduino contains a ``safe/armed" switch along with LED indicators to reduce the risk of accidental injury.

Stem water potential analysis requires the test leaf's stem to be cleanly cut; a damaged specimen would negatively impact the analysis~\cite{scholander1965sap}. Organic matter such as leaf stems exhibit visco-elastic properties. Based on visco-elastic material principles, faster cuts will require less force and result in less deformation of the leaf stem. Our prototype end-effector is able to cut leaf stems with a design target force of $20$\;N at $1.1$\;m/s. The end-effector's chamber has an opening of $110$\;mm by $45$\;mm and a depth of $185$\;mm to accommodate typical avocado leaves. The end-effector is constructed with miniature aluminum extrusions, lightweight 3D printed parts, and laser-cut acrylic panels. The assembly weighs $1.091$\;kg, which is $42$\% of the robotic arm's $2.6$\;kg payload. The end-effector is powered separately from the arm to enable stand-alone testing with a $7.4$\;V 2S LiPo battery.



To determine the types of leaves that can be cut by the mechanism, we performed testing with a variety of trees in a local orchard. The end-effector was manually placed around leaves and activated. Four different crops were selected (avocado, clementine, grapefruit, and lemon) for evaluation. For each crop, ten cutting attempts were performed. Results are shown in Table~\ref{tab:leaf_cutting_agops}. The end-effector was able to cut 95\% of the leaves (38 out of 40). Lower success rates were observed for the lemon and grapefruit leaves. This is due to these particular leaves having shorter stems which made it harder to position the end-effector around the stem without interference from branches or other leaves. The end-effector worked consistently on clementine and avocado leaves.

\begin{table}[!h]
\centering
\caption{Leaf Cutting Tests}
\label{tab:leaf_cutting_agops}
\vspace{-6pt}
\begin{tabular}{cccc}
\toprule
Crop  & Successful Cuts & Attempts & Rate  \\
\midrule
\midrule
Avocado    & 10     & 10                           & 100\% \\
Clementine & 10     & 10                           & 100\% \\
Grapefruit & 9      & 10                           & 90\% \\
Lemon      & 9      & 10                           & 90\% \\
\midrule
Total      & 38     & 40                           & 95\% \\
\bottomrule
\end{tabular}
\vspace{-12pt}
\end{table}


\subsection{Perception}
\label{sec:ta_leaf_perception}

We propose a leaf detection and localization algorithm using 3D point cloud and processed through the Open3D library. Our approach is outlined in Fig.~\ref{fig:systemdiagram}. 
The detection phase seeks to obtain the 3D bounding box of leaves candidates from point cloud captured from the depth camera. First, we remove outliers considered as noise resulting from sensor measurement inaccuracies and segment out the background at a specific distance threshold from the camera frame. Then, downsampling is applied to optimize the performance of the upcoming step. Next, we group the remaining point cloud segments into clusters using the Density Based Spatial Clustering of Applications with Noise (DBSCAN) approach~\cite{Ester1996ADA}. It relies on two parameters, the minimum distance between two points to be considered as neighbors (\textit{eps}) and the number of minimum points to form a cluster (\textit{MinPoints}).

Each resulting cluster is considered a potential leaf and described by a 3D bounding box defined by center $C = [c_x, c_y, c_z]^{T}$, dimensions $D = [h, w, d]$, and orientation $R(\theta, \Phi, \alpha)$. Then, filtering is applied on the clusters using geometric features of the bounding box: number of points, volume, leaf ratio. Finally, the pose of the center of each bounding box is returned as the 6D pose of a potential leaf. 

\begin{figure}[!t]
\vspace{6pt}
    \centering
    \includegraphics[trim={0 0 0 0 },clip,width=0.95\linewidth]{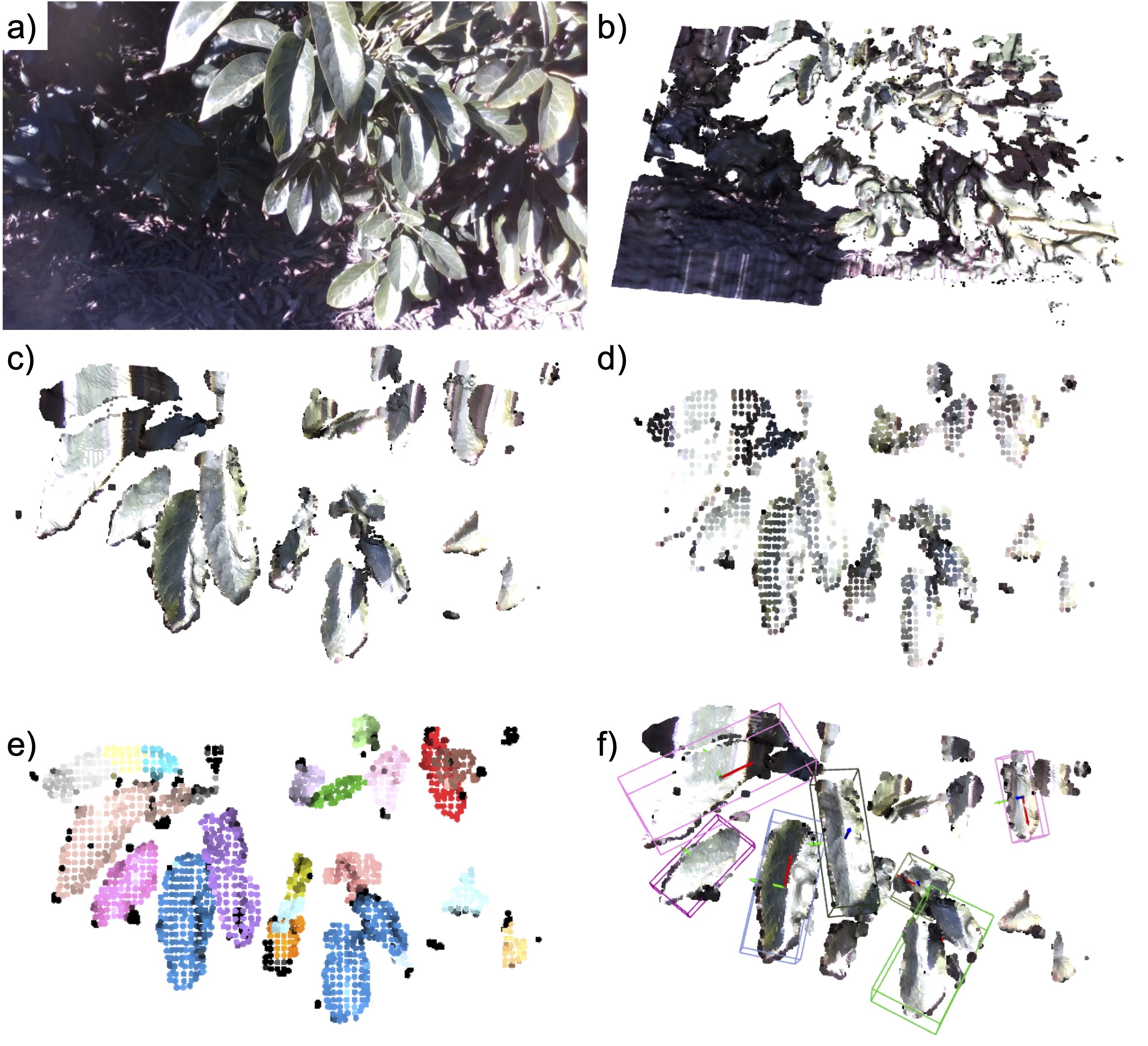}
    \vspace{-12pt}
    \caption{Key steps in our proposed leaf detection and localization process. The sample here corresponds to an outdoor point cloud: (a) corresponding RGB image of the tree, (b) raw point cloud, (c) distance filtered ROI, (d) downsampled point cloud, (e) segmented clusters, and (f) detected candidate leaves without 6D pose bounding boxes.
    }\label{fig:pointcloudoutside}
    \vspace{-18pt}
\end{figure}

To validate our approach, we conducted offline tests for detection and localization separately. For the detection step, ROSbags were collected both in indoor and outdoor settings. 
Indoors (lab with constant light conditions), we used the Kinova arm with the camera placed at different distances ($0.2-0.3$\;m) from a potted tree. 
Outdoors (local orchard with varying light conditions), we collected data manually. We considered a wide range ($0.5-1.6$\;m) of distances from trees; an example is shown in Fig.~\ref{fig:pointcloudoutside}.a. 
A total of 25 point clouds were collected (10 indoor and 15 outdoor). and tested offline with different combinations for \textit{eps} and \textit{MinPoints} parameters, to determine optimal values for later use.

Table~\ref{tab:leaf_pcd_detection} shows the outcome of our experiments on the 10 indoor point clouds and 15 outdoor point clouds.
We attain an average of 80.0\% of detection with a maximum of 90\% for indoor dataset, and an average of 79.8\% with a maximum 85\% for outdoor. 
Further, we observed that the distance between the camera and the tree impacts the optimal values for the point cloud processing. The greater the distance from the camera, the higher \textit{eps} while \textit{MinPoints} decreases. 

\newcolumntype{L}{>{\centering\arraybackslash}m{1.25cm}}
\begin{table}[!ht]\vspace{-6pt}
\centering
\caption{Leaf Point Cloud Detection}
\label{tab:leaf_pcd_detection}
\vspace{-6pt}
\begin{tabular}{LLLLL}
\toprule
 & Point Clouds & Total \# Leaves  & Average Detection & Percentage\\
\midrule
\midrule
Indoor & 10 & 20 & 16 & 80.0\% \\
\midrule
Outdoor & 15 & 99 & 79 & 79.8\%\\
\bottomrule
\end{tabular}
\end{table}

\begin{figure}[!t]
\vspace{6pt}
    \centering
    \includegraphics[trim={0 0 0 0 },clip,width=0.80\linewidth]{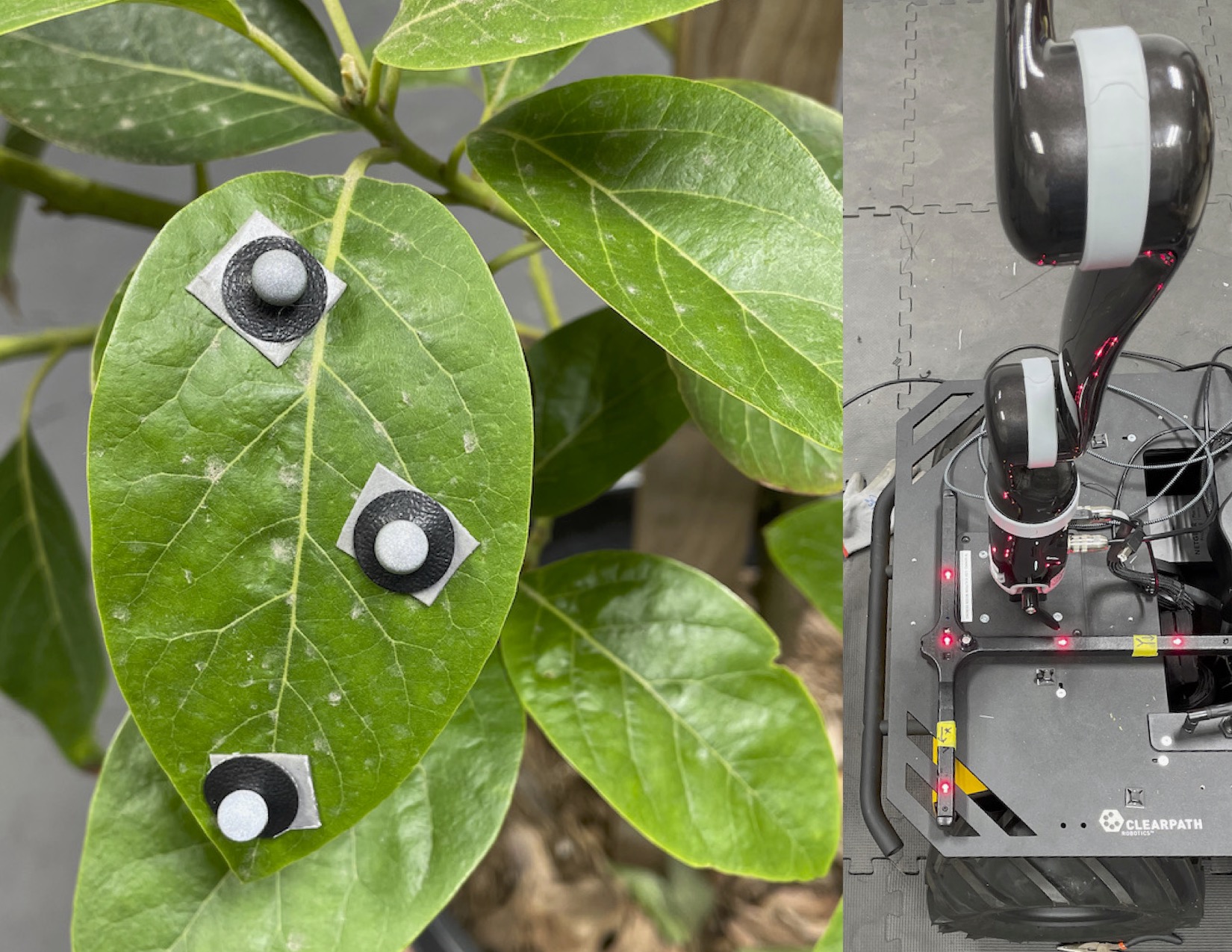}
    \vspace{-6pt}
    \caption{We used motion capture to establish a ground truth for determining the leaf 6D pose. Markers were placed on a target leaf (left) with origin at the base of our 6-DOF robot (right). (A real avocado tree was used.)
    }
    \label{fig:leafvicondots}
    \vspace{-18pt}
\end{figure}

To validate the localization phase, we compare several 6D poses obtained via our proposed approach against ground truth data obtained from a VICON motion capture camera system. Retroreflective markers were placed around the center of leaves, as shown in Fig.~\ref{fig:leafvicondots}, to estimate their pose.

Table~\ref{tab:leaf_pose_error} summarizes the results obtained for 12 random leaves positions. Our approach provides an estimation with mean error of 8.28 mm, 14.38 mm, and 15.54 mm along x-axis, y-axis, and z-axis, respectively,for avocado leaves of width ranging between $24-86$\;mm and length ranging between $54-150$\;mm. Based on the average leaf size ($48\times91$\;mm), estimation errors represent nearly 15\% of the width and 17\% of the length. We evaluated the orientation by calculating the Euclidean distance between the two provided values using the definition in~\cite{Huynh2009MetricsF3}. We obtained a mean error of 5.3$\deg$.   
We observe that the obtained 6D pose may drift from the physical center of the leaf mainly on the y-axis and z-axis due to human-induced error and the non-rigid nature of the leaf which impacts marker placement.

\begin{table}[!ht]
\centering
\caption{Leaf 6D Pose Error}
\label{tab:leaf_pose_error}
\vspace{-6pt}
\begin{tabular}{ccccccc}
\toprule
 Error & $\Delta x$ (mm)& $\Delta y$ (mm) & $\Delta z$ (mm)& Orientation ($\deg$)  \\
\midrule
\midrule
Mean & 8.28 & 14.38 & 15.54 & 5.3\\
\midrule
Std dev & 7.46 & 5.46 & 6.69 & 15.5 \\
\bottomrule
\end{tabular}
\end{table}


The proposed approach provides an initial 6D pose along useful information of potential leaves using a processed 3D point cloud and obtained up to 80\% of detection and a mean error less than 16mm and 5.3 $\deg$. Both detection and localization steps were performed without the need of collection or storage of large data including 3D models, and training process. Furthermore, all tests were run using a CPU configuration, without any additional GPU acceleration.

\section{Experimental Validation of Leaf Cutting}
\label{sec:exp_methods}



\begin{figure*}[!t]
\vspace{6pt}
    \centering
    \includegraphics[trim={0cm 0cm 0cm 0cm },clip,width=0.95\linewidth]{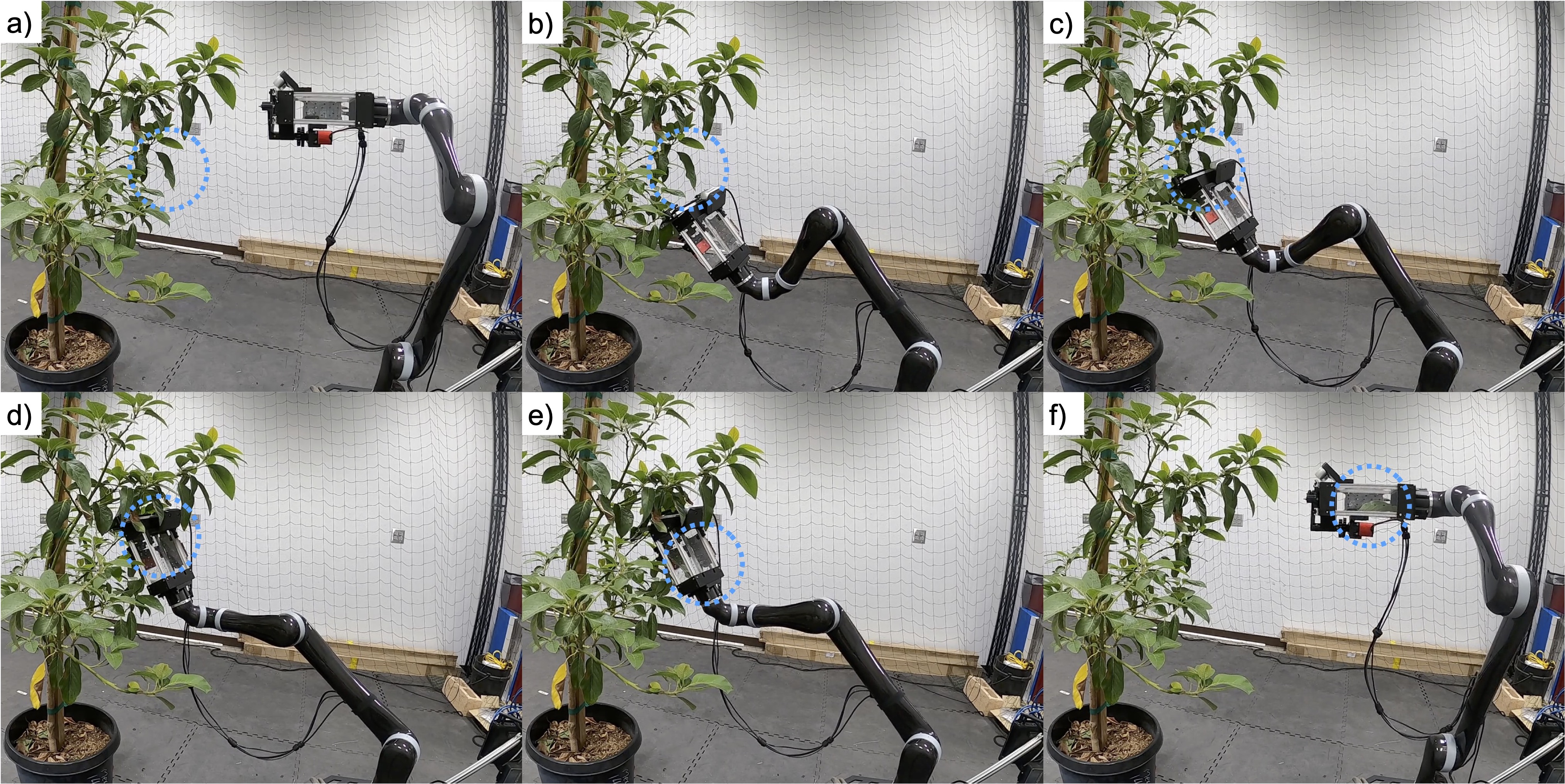}
    \vspace{-6pt}
    \caption{Overall leaf retrieval process. During the perception phase, (a) the point cloud is processed to determine a potential leaf. If a viable leaf is detected, (b) the arm will move to an offset position. (c) The arm will then perform a linear motion to capture the leaf. Once in position, (d) the arm will cut the leaf and (e) the leaf will fall into the enclosed chamber. (f) After completing the cut, the arm will return to the home position.
    }
    \label{fig:trialsequence}
    \vspace{-18pt}
\end{figure*}

To evaluate our overall leaf detection, localization and cutting pipeline, we tested with a real potted avocado tree indoors (lab). The mobile manipulator and end-effector system was positioned at random poses near the base of the tree so that the end-effector was at distances ranging between $0.2-0.3$\;m from the edge of the tree canopy. An experimental trial consisted of collecting a point cloud, storing the identified and localized potential leaves in a queue, and then sending the queued leaves to the arm for a retrieval attempt. Each trial concluded once the queue was depleted and the tree was repositioned for the next trial. Figure~\ref{fig:trialsequence} outlines this process.

For each retrieval attempt, leaf candidates and viable leaves are determined. \emph{Leaf candidates} are leaves that have a pose within the arm's workspace. \emph{Viable leaves} are leaf candidates that have a retrieval path within the arm's workspace. For testing our point cloud detection, we are interested in monitoring both successful captures and successful cuts of the leaf. A \emph{successful capture} occurs when the end-effector is placed around a viable leaf while a \emph{successful cut} occurs when the enclosed leaf is removed from the tree. A \emph{clean cut} occurs when the leaf is severed cleanly at the stem such that it could be used for stem water potential analysis.

Out of 46 trials, 63 potential leaves were detected by the point cloud.  (Note that each point cloud in the trial could produce a variable amount of leaves, hence a higher number of potential leaves than trials.) After filtering the potential leaves to remove the leaves outside of the work space, 39 viable leaves remained. Out of these leaves, 27 were captured successfully (69.2\%) while 21 of the 27 captured leaves were cut (77.8\%). Table~\ref{tab:succ_rates} summarizes retrieval results while Table~\ref{tab:proc_time} highlights the process times. The mean point cloud processing (perception) time was $5.6$\;sec and the mean cutting (actuation) time was $10.6$\;sec. The mean total retrieval time was $16.2$\;sec.

\begin{table}[!ht]
\centering
\caption{Leaf Retrieval Numbers \& Rates}
\label{tab:succ_rates}
\vspace{-6pt}
\begin{tabular}{lrr}
\toprule
Stage               & \multicolumn{1}{l}{Number} & \multicolumn{1}{l}{Rate} \\
\midrule
\midrule
Potential Leaves    & 63                         & \multicolumn{1}{l}{N/A}  \\
Candidate Leaves    & 51                         & 81.0\%                   \\
Viable Leaves       & 39                         & 76.5\%                   \\
Successful Captures & 27                         & 69.2\%                   \\
Successful Cuts     & 21                         & 77.8\%                   \\
\midrule
Clean Cuts          & 4                          & 19.0\%                   \\
Near Misses         & 7                          & 30.0\%                   \\
\bottomrule
\end{tabular}
\end{table}

\begin{figure}[!t]
\vspace{6pt}
    \centering
    \includegraphics[trim={0 0 0 0 },clip,width=0.8\linewidth]{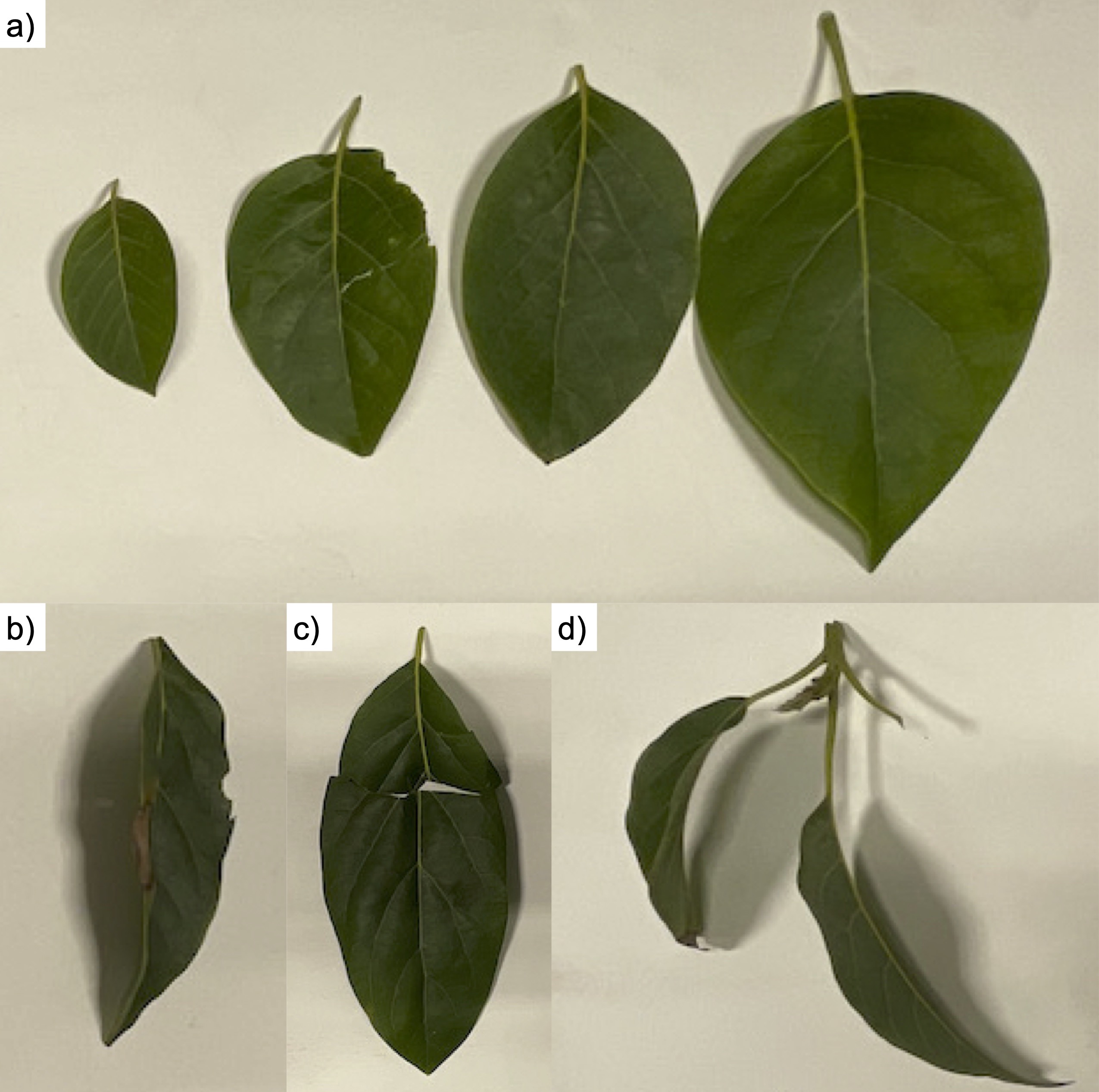}
    \vspace{-6pt}
    \caption{Sample leaves cut from our avocado tree during automated indoor tests. (a) The four leaves represent clean cuts suitable for stem water potential analysis. (b) The system also cut seven leaves that were classified as near-misses, which removed the leaf without the stem. (c) The remaining leaves were cut closer to the center, due to interference between the end-effector and the branches. (d) In two cases, collateral damage occurred when a second leaf was removed along with the target leaf. These instances were classified as a single successful cut, but not a clean cut since the two leaves would need to be separated for stem water potential analysis.
    }
    \label{fig:leafcutarray}
    \vspace{-18pt}
\end{figure}

Our system was able to remove a total of 21 leaves from the tree. However, not all leaves were clean cuts on the stem; four were classified as clean cuts for use in stem water potential analysis. The majority of the leaves were severed at the top of the leaf and not at the stem (Fig.~\ref{fig:leafcutarray}). Our system produced seven near-misses where the leaf was cut within an average of $9.58$\;mm from the stem (std dev: $6.1$\;mm). The remaining 10 leaves were severed closer to the middle of the leaf, largely due to collisions with the branches. Similar branch interference also lead to four out of the six missed cuts from the captured leaf. These two problems could be solved in future work through a refined end-effector design, more robust path planning to account for branches, and implementing visual servoing for continuous stem alignment as the end-effector approaches a viable leaf.

\begin{table}[!ht]
\centering
\caption{Leaf Retrieval Performance Time (Seconds)}
\label{tab:proc_time}
\vspace{-6pt}
\begin{tabular}{lrrr}
\toprule
Metric  & Perception Part & Actuation Part & Overall Retrieval \\
\midrule
\midrule
Min  & 0.5                         & 4.6                       & 6.1                     \\
Max & 11.0                       & 61.7                      & 62.5                    \\
Mean & 5.6                         & 10.6                      & 16.2                    \\
Median  & 7.7                         & 8.1                       & 15.3                    \\
Std dev  & 3.9                         & 10.4                      & 10.2  \\      
\bottomrule
\end{tabular}
\end{table}

\section{Conclusions}

Our work develops a co-designed actuation and perception method for leaf identification, 6D pose estimation and cutting. Our developed leaf-cutting end-effector can cut leaves of various types of trees (avocado, clementine, grapefruit and lemon) cleanly at their stem with a 95\% success rate on average. Our proposed 3D point cloud technique can be successful for detecting an average of 80.0\% of leaves indoors and 79.8\% outdoors, and localizing them with less than 17\% error along the leaf's length or width. 
Experimental testing of the overall proposed framework for leaf cutting reveals that our system can capture 69.2\% of viable leaves and cut 77.8\% of those captured leaves. 

These results offers a promising initial step toward automated stem water potential analysis, nonetheless several steps remain and are exciting avenues for future work. 
The end-effector can effectively cut the leaves, but its size presents a challenge when cutting certain leaves like those from lemon and grapefruit trees which in turn calls for further design optimization. The current path planning approach works well for leaves that are on the periphery of the tree's canopy. Alternate path planning strategies can be explored to reach leaves within the canopy closer to the trunk, and integrated with visual servoing to better align the cutter with the stem of the leaf as it is about to cut it.
Furthermore, the system will need to be robust to disturbances such as wind before deployment in an outdoor orchard environment.
Finally, to enable automated stem water potential analysis, the captured leaf will need to be transferred from the end-effector into a pressure chamber.

\bibliographystyle{IEEEtran}
\bibliography{Amel_bibliography,Kostas_bib_from_icra,Merrick_bib_new}

\end{document}